\documentclass[journal]{IEEEtran}
\usepackage{graphicx}
\usepackage{linguex}
\usepackage[utf8]{inputenc}

\usepackage{natbib}
\bibpunct{(}{)}{;}{a}{,}{;}
\usepackage{xcolor}
\usepackage[many]{tcolorbox}

\newfont{\eightss}{cmssq8}                      % small sans serif
\usepackage{fancyhdr}
\title{How Lexical Gold Standards Have Effects On The Usefulness Of Text Analysis Tools For Digital Scholarship}
\author{Jussi Karlgren \\
KTH Royal Institute of Technology and Gavagai, Stockholm}
\begin{document}
\maketitle

\begin{tcolorbox}[colback=red!10!white,
                     colframe=red!20!black,
                     title=\textsc{A shorter version of this paper has been published at CLEF},  
                     center, 
                     valign=top, 
                     halign=left,
                     before skip=0.8cm, 
                     after skip=1.2cm,
                     center title, 
                     width=3in]

  Karlgren, Jussi. "How lexical gold standards have effects on the usefulness of text analysis tools for digital scholarship." In Proceedings from {\em International Conference of the Cross-Language Evaluation Forum for European Languages}. Springer, 2019.

  \end{tcolorbox}

\begin{abstract} % 150 - 250 wds
This paper describes how the current lexical similarity and analogy gold standards are built to conform to certain ideas about what the models they are designed to evaluate are used for. Topical relevance has always been the most important target notion for information access tools and related language technology technologies, and while this has proven a useful starting point for much of what information technology is used for, it does not always align well with other uses to which technologies are being put, most notably use cases from digital scholarship in the humanities or social sciences. This paper argues for more systematic formulation of requirements from the digital humanities and social sciences and more explicit description of the assumptions underlying model design.

% \keywords{Gold standard  \and Lexical semantics  \and Digital scholarship.}
\end{abstract}

% Capitalise section headers
\section{Text Analysis Is Mostly Based On Lexical Features}
Text analysis technology is almost exclusively based on lexical features, i.e. on observing the presence and the frequency of occurrence of words in a text or a section of text of interest. These observations are used typically for classifying or scoring texts. Features are treated variously by algorithms ranging from simple observation of presence, to frequency calculations, or to non-linear combinations of items using e.g. neurally inspired models. In most cases, algorithms rely on background lexical models primed by observations made on large amounts of data to be able to discern what lexical features are of specific interest in the data set at hand. % Without knowing that X and Y are Z we cannot know if the presence of X in text is a notable observation or a humdrum one. 

Evaluating the quality of such background models is made using semantic tests of some generality, intended to provide a reasonable sample of language to capture the general competence of a model.

\section{Predication of Aboutness}
Human linguistic behaviour rests on \textit{predications}: a speaker or author indicates some referents of interest and formulates something of interest about them, relating them to each other or to preceding discourse. Prototypically, referents are noun phrases; the relations among referents, between referents and the discourse itself, or between referents and the surrounding context are prototypically verb phrases. This admittedly very simplified model of how semantics and pragmatics work in functional discourse will serve to elucidate some challenges for evaluating lexical resources given below.

\section{Referential Semantics and Topicality}
Referentiality covers one of the more important aspects of language use: that of topicality, where language calls up items, concepts, notions of interest to discourse participants. In most computational text analysis tasks, topicality has been at the center of attention: what a text is \textit{about} is the primary categorisation criterion. The general intuition of topical analysis is that many terms in language appear in a tight bursty pattern to indicate that some matter of interest is under treatment, and that other terms appear in a wider distribution, constituting structural material rather than topical. As an example, texts which contain terms \textit{helicopter}, \textit{rotor}, \textit{airfield}, and \textit{pilot} vs texts which contain the terms \textit{cow}, \textit{milk}, \textit{dairy}, and \textit{barn} can with some ease be classified topically from bursty term occurrence alone, finding terms that are unexpectedly common compared to language usage in general. Terms such as \textit{see}, \textit{move}, \textit{rotate}, or \textit{yield} are not as useful for this purpose. This leads to quality criteria for text analysis tools related to \textit{coverage}: to ensure recall, a classifier must be able to find semantically related terms, if some initial terms have been given. These may be synonyms or near synonyms (\textit{autogiro, chopper, whirlybird}) or other related terms (\textit{airfoil, camber, translational lift}). % distributional models trained on a close context (Sahlgren et al) will give synonyms and other lexicosemantic relations. these will be the same word class. 

\section{Non-topical, Less Referential Semantics}

Much of what is in a text does not directly contribute to its topicality. The text also organises the structure of the discourse into appropriately complex chunks, aids the listener or reader to achieve coherence in what is being communicated, indicates speaker or author attitude and stance, and communicates temporal and process qualities of the predication given. A text also evokes other texts and other usages through its stylistic and lexical choices, by adhering to conventions, by quoting, paraphrasing, or reformulating other works and other authors. Some such qualities of the text are highly rule-bound and conventionalised, others are free for the author or speaker to make explicit if they should so wish. Some such qualities are general over an entire discourse, with observable surface items sprinkled throughout the textual data and thus cannot be pinpointed to any single utterance or to the occurrence patterns of some small set of linguistic items.  

There is no obvious and crisp definitional demarcation between referential and topical variation and more general thematic or attitudinal variation: on an operational level they vary between such linguistic items with localised occurrence patterns in text and such that permeate the entire body of text under consideration.

\section{Close and Distant Reading}
In recent years, research in the humanities has adopted the possibility of working with collections of documents rather than small focussed selected sets. The attendant methodological debate is frequently framed as a distinction between {\em close reading}, the traditional approach of the humanities to \textit{engage} closely with cultural items---in this case, texts and their contexts---and \textit{distant reading}, the potentially fruitful set of methods having to do with working on comprehensive data sets from e.g. a certain period, genre, or class of author, using computational tools, visualisation and graphing techniques, and overview analyses to find patterns which would not have been notable using traditional methods \cite{moretti2013distant}.

These new tools, new methods, and new results are not universally welcomed by scholars in the humanities. The debate over how to best use new technologies is lively and goes to the roots of what the ultimate research goals of the humanities and the social sciences are. The humanities and the social sciences do not only have different methods than engineering and the natural sciences do, but their goals and aims when they produce knowledge are different, and they approach information differently 
\cite{fitzpatrick2011humanitiesdonedigitally,da2019debacle,underwood2019dearhumanists,da2019computationalcase}. Debate notwithstanding, it is not difficult to compile a long and comprehensive list of research questions, most of which are only incidentally topical in nature: how authors and schools of thought spread and influence each other, how much or little knowledge of distant cultures there was at some time in some cultural area, how political institutions change over time, how argumentation influences decision making, how public sentiment affects financial indicators, how the well-being of individuals are manifested in their writing, how a scholarly field selects its focus topics, how language change is motivated by local prestige markers, how social change is reflected in literary work, how to determine who has authored a given work, and so forth \cite[e.g.]{janicke2015closeanddistantreading,oconnor2011computationaltextanalysisforsocialsciences}. Many of these questions are supremely amenable to large scale work on collections, even when they are only incidentally topical in nature, and many of these questions have been touched upon or addressed directly in recent years in experimental work here at CLEF. 

From the point of view of information access research, we can expect quite interesting new use cases to emerge for language technologists and information access researchers to work with once the methodological discussions in the humanities settle: the topical content of texts or text is only one of the objects of study, engagement in the material is the prime method, and future computational tools will be there to allow for new types of engagement in more extensive collections of material. 

\section{Difficult and Simple Tasks}

It is worth noting that in the general processing and analysis of human-generated data there are simple tasks where the challenge is about scale and consistency, not interpretation: information retrieval, e.g. A rational way to approach such tasks is to simplify the collection thereby reducing the variation, e.g. by reducing texts to bags of words. There are also difficult tasks, where humans struggle to extract information of interest reliably from data: authorship attribution, novelty detection, textual entailment, trustworthiness of text e.g. Here, the task of computational analysis is not address scale, but to help uncover and evaluate the effectiveness of features or feature combinations which are difficult to discern for a human analyst. There is no reason to assume that the same computational approaches are effective for both classes of challenge!

Some of the tasks under consideration in digital scholarships can well be categorised as difficult tasks. 

\section{Target Notions for Language Technology and Text Analysis}
So what effects do the observations given above have for evaluation of information systems? Most tools built for information access have explicitly stated goals to optimise for topical relevance, for timeliness to fulfil some typically current information need on the part of the user. This goes together well with referential semantics. In view of the preceding discussion on referentiality this translates to observable and computable \textit{burstiness} as an attractive operational target notion to decide which items mentioned in a text are useful to characterise it \cite{katz1996distribution}. 

For this reason, one of the very effective mechanisms in document processing is that of term weighting. The idea behind term weighting is selectivity: what makes a term valuable is whether it can pick any of the few relevant documents from the many non-relevant ones. Karen Spärck Jones defined what was to become the \textit{idf} measure in 1972: \textit{"It is argued that terms should be weighted according to collection frequency, so that matches on less frequent, more specific, terms are of greater value than matches on frequent terms"} \cite{sparck1972statistical} and this measure has been adopted---for good reason---in just about every term weighting mechanism in use today. This measure weights terms according to their topical specificity: how well they distinguish documents from each other by way of referential content.\footnote{Spärck Jones argues that this should not be understood in terms of semantics, but in terms of occurrence statistics, but the target notion is a relevance-oriented one.} This is a sensible approach if topical relevance is the target notion. 

\setlength{\tabcolsep}{15pt}
\begin{table*}[htbp]
\caption{Lexical categories and their occurrence frequencies in documents.}\label{pos}
          \begin{tabular}{r|rrr}
            &           \textit{nouns} &    \textit{adjectives}      &         \textit{verbs} \\
            \hline
number of occurrences            &  24 000 000 & 5 000 000 & 7 000 000 \\ % 24349780 5200970 6730550
number of different items            &     200 000 &    25 000 &     8 000 \\ % 210797 24494 7656
            \hline
$>  100$ documents   &          6\% &      15\% &      28\% \\  % 11835 3675 2137
$>  200$ documents   &          4\% &       9\% &      20\% \\ % 7912 2322 1537  
$> 1000$ documents   &          1\% &       3\% &       8\% \\  % 2641 696 595
\hline
     \multicolumn{4}{c}{more than twice in the same document} \\
\hline
$>  100$ documents   &          2\% &       3\% &       6\% \\
$>  200$ documents   &          1\% &       2\% &       4\% \\
$> 1000$ documents   &        0.4\% &     0.5\% &     1.5\% \\
          \end{tabular}
 \end{table*}
% 170 255 documents, 5726822 utterances, 72339348 words

\section{Lexical Categories and Occurrence Statistics}

Table~\ref{pos} gives some observed statistics, computed over a collection of two years of news text, with 170~000 documents and more than 72 million words. We find here---as expected---that there are many more noun occurrences than adjectives or verbs, and that those occurrences come from a much larger lexicon of nouns: 200~000 nouns occur more than 24~000~000 times in the materal as compared to 8~000 verbs occurring 7~000~000 times. We find that only a small proportion of the nouns occur in more than 100 or 1000 of the documents, and even less if we count the number of documents they occur more than twice in; the distribution of verbs and adjectives is very different. Understanding these observations in terms referential semantics we can posit in drastically simplified terms that language, using noun phrases, can refer to an unimaginably wide variety of entities, and, using verbs, to a more constrained variety of events or processes.

\section{What Do Lexical Gold Standards Look Like?}
There are several test sets that are used specifically for experimentation with how choice of representation, algorithm, and training set jointly contribute to the qualities of a semantic model. Most, as can be seen in Table~\ref{tab1}, are focussed on nouns and relations between nominals. This, given the discussion above, is unsurprising. There several further experiment sets with a broader selection of lexical classes, but they tend to be embedded into more specialised conceptual models, such as semantic role labeling, word sense disambiguation, or other pragmatic constraints, which raise the threshold for including them in a standard test. These standard test sets have been proven to be quite useful tools to develop lexical resources, as can be seen from their widespread adoption in various evaluation experiments. However, even if a tool built on top of them professes to be general and use case independent, the background lexical resources will have to some extened tuned the tool to fit these ostentatively general gold standards, which in turn will have tool implicitly yield results optimised for topical analysis.

% http://u.cs.biu.ac.il/~dagan/publications/directional-distsim.pdf

%also wikipedia headers and dbpedia/yago etc status

\section{A Case In Point: Topic Models}
One point raised in the methodological debates referred to above is that digital scholarship frequently is driven by "prosaic explanations or choices baked into the method" \cite{da2019debacle}. As a case in point, in practical application of computational tools to digital collections, a frequent approach is to apply \textit{topic models} in various more or less standard formulations to some collection of interest. Some discussion of parameter settings and modularity and on which preprocessing steps are useful and which are destructive can be found---e.g. \cite{tangherlini2013trawling}---but it is taken as a given that topic oriented toolkits are a useful tool to work with and much of the discussion in the humanities today centres on \textit{visualisation} rather than what the text analysis algorithms do with the materials they process \cite{janicke2015closeanddistantreading}. Underscoring this is the observation that models have various requirements on how the incoming data are fashioned and this requires the data to be preprocessed accordingly.

\begin{quote}
    "We then remove stop-words and transform each segment into a vector to compute the similarity between every pair of segments."
\end{quote}

\noindent
Virtually every How-To and instruction for newcomers to text analysis in general, to topic modelling, or other text clustering activities recommends "stopword removal" from texts, most mention punctuation removal, many include stemming or lemmatisation as useful feature reduction steps, and in many cases various additional statistical filters to remove unusual words. These all reduce the complexity of the linguistic signal to facilitate processing in further steps. An unspoken assumption in most cases is that "feature" is taken to be a synonym of "word". Again, this is a sensible approach if topical relevance is the target notion. For other purposes, the feature reduction may effectively remove exactly those features that best capture the variation of interest. Typical output of topic models are then presented either as lists of topics with strongly loaded terms as descriptors or even word clouds. These may or may not give insights, but almost certainly those insights will not be in tune with what previous practice has worked with, which makes the worth of advancements in such directions difficult to assess.

The arguments givene above apply to most similar models, including end-to-end classifiers: if they are built to attend to burstiness, they will search out primarily bursty features and struggle to reconcile it with target notions that would be better modeled in some other way.

\setlength{\tabcolsep}{5pt}

\begin{table*}[htbp]
\caption{Some typical test sets for lexical similarity.}\label{tab1}
\begin{tabular}{|l|r|rrrr|lr|}
\hline
Test sets &  Size & Nouns or NP & Adjectives & Verbs & Other & Reference & Year \\
\hline
RG & 65 & 100\%    &       &       &       & \cite{rubenstein1965contextual} & 1965 \\
Chiarello et al & 144 & 100\%    &       &       &       & \cite{chiarello1990semantic} & 1990 \\

TOEFL      &    80   &    21\%   &  25\%     &  21\%     &  32\%     &  \cite{landauer1997solution} & 1997 \\
WordSimilarity        & 353      &   97\%    &  1\%     &  2\%     &       &  \cite{finkelstein2001placingsearchincontext} & 2001 \\
%Lexical Inference        & 2~202~568      &   76\%    &       &  24\%     &       &  \cite{kotlerman2010directional} & 2010 \\
ConceptSim       &       &   100\%    &       &       &       &   \cite{schwartz2011evaluating} & 2011 \\
BLESS       &   200    &  100\%     &       &       &       &   \cite{baroni2011bless} & 2011 \\
Entailment       &   15~992    &  100\%     &       &       &       &   \cite{baroni2012entailment} & 2012 \\
Syntactic Analogy        &   8~000    &  25\%     &  37.5\%     &   37.5\%    &       &   \cite{mikolov2013linguistic} & 2013 \\
SIMLEX        &   999    &  67\%     &  11\%     &   22\%    &       &   \cite{hill2016simlex} & 2016 \\
\hline
\end{tabular}
\end{table*}

\section{Lessons Learned and Paths Forward}
In conclusion, it is crucial to understand that gold standards are not use case free, not even intrinsic ones such as term lists, and that applying them to optimise technologies will have downstream effects. The effects of the implicit use cases in evaluation are of varying importance. If tools are used for a slightly different purpose from what they are designed for, everything may work out perfectly fine; sometimes it may not. The major risks are that technologies may obscure that what most interests its users in digital scholarship, which will first result in shoddy or uninteresting research, with an attendant backlash and skepticism to computational methods in general. Both of these effects are evident already. 

Much of the intradisciplinary debate in the various fields of digital scholarship is based on a prejudicial view of what the aims of engineering and the natural sciences are: "the sciences simplify, where the humanities embrace complexity". These sort of statements are frequently accompanied by calls for engineers to study more humanities. While this is a worthy goal in general and might make engineers happier people, it would not necessarily improve the tools used in digital scholarship. More important is for those who use tools to examine what the tools are built to work with and to take responsibility of those assumptions when they draw conclusions from their output. If those assumptions fit poorly with the tasks they intend to address, they should request other tools. 

% That understanding is something that should be supplied by those who build the systems, and this means that system builders and algorithm designers must be attentive to the interests and needs as expressed by those who use the tools, especially when they are applied in ways which were not quite the expected.

Similarly, those who design, build, and evaluate tools to pay more attention to what underlying assumptions they bring in with the technology components and evaluations procedures they include, and to make them known to those who use the tools further on down the line. There are systematic use case description frameworks which are useful for these types of crossdisciplinary bridging. 

For those of us who worry about systematic evaluation, we must make sure to engineer gold standards to ensure their coverage over a larger space of downstream use cases, and to document the underlying assumptions and observable distributional characteristics of the gold standard items in greater detail. This applies to simple lexical gold standards discussed in this paper, but also to more sophisticated sets of texts and utterances. An additional specific point to address here is that feature engineering a gold standard to be used for evaluating a technology component cannot fairly be done by those who build the technology, but neither can we expect the typical end user to be able to do so, since much of the feature space under consideration is opaque for engineers and humanities scholars or social scientists alike. Building a fair and reasonably representative gold standard for text will involve analysis of the character of textual material in general and the items under consideration specifically. This requires both an understanding of the feature space and an ability to describe it formally. Obviously, other modalities will involve similar concerns. This means that the construction of gold standards should involve expertise in the material the gold standard is fashioned from: for text, linguists or philologists; for images, art analysts, historians, or practitioners, and so forth.

%If the only tool we offer is a hammer we are sort of implying that all problems are loose nails.

\bibliographystyle{plainnat}
{\small
  \bibliography{facit}
  }
\end{document}